\documentclass[a4paper,conference]{IEEEtran}
\usepackage{setspace}
\usepackage{pbox}
\usepackage{array}
\usepackage{wrapfig}
\usepackage{times}
\usepackage{epsfig}
\usepackage{graphicx}
\usepackage{amsmath}
\usepackage{amssymb}
\usepackage{array}
\usepackage{soul}
\usepackage{xcolor}
\usepackage{url}
\usepackage{multirow}
\usepackage{etoolbox,lineno}
\usepackage{commath}
\usepackage{soul}

\ifCLASSINFOpdf
\else
\fi
\begin{document}
%
\title{Super-resolution Guided Pore Detection for Fingerprint Recognition}



%
\author{\IEEEauthorblockN{Syeda Nyma Ferdous\IEEEauthorrefmark{1},
Ali Dabouei\IEEEauthorrefmark{1},
Jeremy Dawson\IEEEauthorrefmark{1} and
Nasser M Nasrabadi\IEEEauthorrefmark{1}}
\IEEEauthorblockA{\\
Lane Department of Computer Science and Electrical Engineering, West Virginia University, USA\IEEEauthorrefmark{1}\\  \textit{\{sf0070, ad0046\}@mix.wvu.edu,  \{jeremy.dawson,nasser.nasrabadi\}@mail.wvu.edu}}}


\maketitle

\begin{abstract}
Performance of fingerprint recognition algorithms substantially rely on fine features extracted from fingerprints. Apart from minutiae and ridge patterns, pore features have proven to be usable for fingerprint recognition. Although features from minutiae and ridge patterns are quite attainable from low-resolution images, using pore features is practical only if the fingerprint image is of high resolution which necessitates a model that enhances the image quality of the conventional 500 ppi legacy fingerprints preserving the fine details. To find a solution for recovering pore information from low-resolution fingerprints, we adopt a joint learning-based approach that combines both super-resolution and pore detection networks. Our modified single image Super-Resolution Generative Adversarial Network (SRGAN) framework helps to reliably reconstruct high-resolution fingerprint samples from low-resolution ones assisting the pore detection network to identify pores with a high accuracy. The network jointly learns a distinctive feature representation from a real low-resolution fingerprint sample and successfully synthesizes a high-resolution sample from it. To add discriminative information and uniqueness for all the subjects, we have integrated features extracted from a deep fingerprint verifier with the SRGAN quality discriminator. We also add ridge reconstruction loss, utilizing ridge patterns to make the best use of extracted features. Our proposed method solves the recognition problem by improving the quality of fingerprint images. High recognition accuracy of the synthesized samples that is close to the accuracy achieved using the original high-resolution images validate the effectiveness of our proposed model.
\end{abstract}


%
\IEEEpeerreviewmaketitle

\section{Introduction}

Fingerprint is one of the principal biometric traits which has been widely adopted for person recognition due to the highly distinctive nature and simplicity of acquisition \cite{maltoni2009handbook}. The success of using fingerprints in different biometric applications, such as forensics and access control, has led to the intensive development of fingerprint recognition systems. Feature extraction is a major component of fingerprint recognition that directly impacts the matching performance. Fingerprint features can be broadly categorized into three levels. Level-1 features include the coarse and global structure of the ridge pattern, and therefore, contain low discriminative information. Level-2 features are  called  minutiae and are the most widely used type of features in fingerprints. Minutiae are specific patterns within the ridges, mainly formed by ridge ending and ridge bifurcation. Level-3 features represent the fine attributes that are rich in quantitative information prevalent among sweat pores in fingers that could be leveraged for high-accuracy identification. 

Several studies have demonstrated that level-3 features can significantly improve the matching performance of fingerprint recognition systems \cite{jain2006pores, su2017deep, anand2020porenet}. However, employing level-3 features imposes challenges that can limit the efficiency of the matching system. Level-1 and level-2 features can be reliably extracted from low-resolution images generally captured at 500 ppi or less, while level-3 features demand capturing high-resolution fingerprints at resolution above 700 ppi \cite{zhang2010selecting}. This negatively affects the cost of the device since the sensor must be able to capture detailed fingerprints. Furthermore, finger pores are small in size at low resolution, thus conveying much less information for a feature extractor to ensure effective recognition performance. In addition, pore information cannot be effectively extracted from the legacy fingerprints captured at the conventional 500 ppi resolutions. 

To overcome these challenges, we evaluate the feasibility of extracting level-3 features from low resolution fingerprints using a learning-based framework guided by Super-Resolution (SR); a technique where a High-Resolution (HR) fingerprint image is generated from a Low-Resolution (LR) fingerprint image. Reconstruction of high-resolution fingerprints from low details makes this scheme more challenging. Recently the relationship between super-resolution and object recognition has been studied in several works \cite{haris2018task,shermeyer2019effects,ferdous2019super} delineating the effect of super-resolution on object recognition performance. All of these studies point to improvement of recognition performance as SR techniques allow a more rigorous analysis of features being used by a detector.  

\begin{figure}[b]
\centering 
 \includegraphics[width=8cm,height=3.3cm]{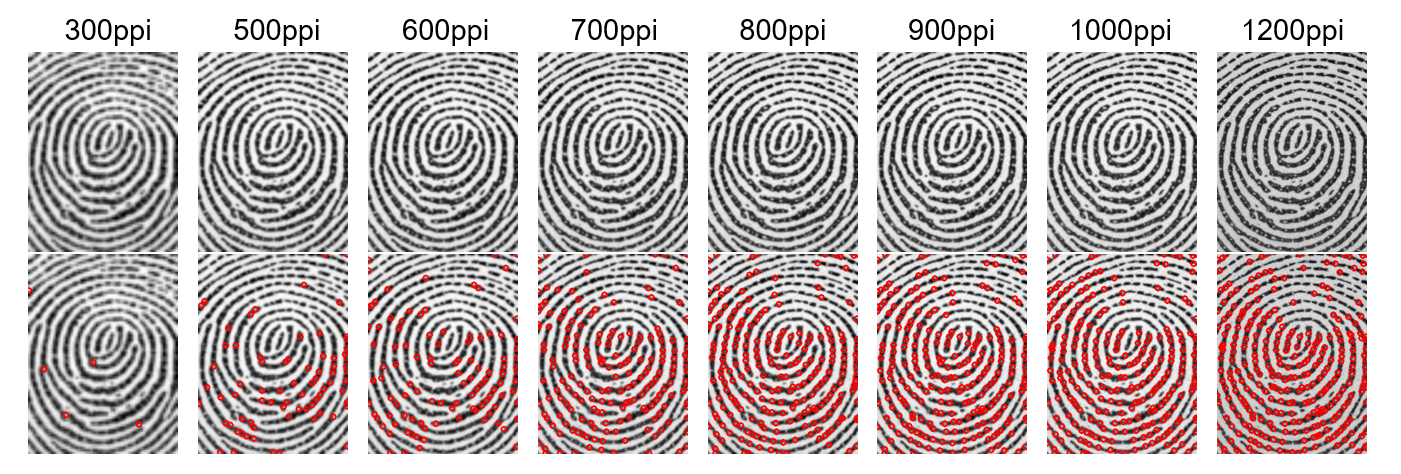}
  \caption{\label{fig:reso} Visual representation of pore detection performance at varying resolution. Red circles represent the detected pores. An increase in the number of identified pores is apparent with the increase in resolution.}
\end{figure}
Pore detection is a crucial step in designing a fingerprint recognition system which greatly impacts the overall system performance. A remarkable pore detection accuracy is achievable with an increase in resolution as shown in Fig. \ref{fig:reso}. Therefore, we propose a modified single-image SR algorithm using the Super-Resolution Generative Adversarial Network (SRGAN) \cite{ledig2017photo} tailored for fingerprint pore detection. We have used all three level features namely minutia, ridge pattern and pore features extracted from fingerprints for better recognition performance. The main contributions of this paper are as follows:

\begin{itemize}
  \item We develop a deep fingerprint SR model which employs SRGAN to reliably reconstruct high resolution fingerprint samples from their corresponding low-resolution samples.
  \item We adopt a pore detection scheme that helps the SRGAN model to focus on level-3 features while synthesizing HR fingerprint samples. A jointly trained deep SR and pore detection framework is proposed. 
  \item To better utilize the ridge information of fingerprint samples in combination with pores, we have incorporated a ridge reconstruction loss making use of level-2 and level-3 features in our overall objective function, which helps to improve the fingerprint recognition accuracy of our model. 
   \item In addition, to make sure that the framework retains class identity, we have used an auxiliary deep verifier module combined with a quality discriminator to conduct fusion at the feature level. 
\end{itemize}

\section{Related Work}
\subsection{Fingerprint Recognition Using Pores}
In recent years, extensive research has been conducted on fingerprint matching utilizing pore information. Early studies \cite{stosz1994automated, kryszczuk2008extraction} mainly followed skeleton tracing approaches for pore detection. Stosz et al. \cite{stosz1994automated} first used a multi-level fingerprint matching approach employing pore position and ridge feature information in skeletonized fingerprint samples. Apart from the skeletonization approach, filtering based methods have also been applied for pore extraction. Jain et al. \cite{jain2006pores} developed a fingerprint matching algorithm making use of a Mexican hat wavelet transform and Gabor filters for automatic extraction of pores and ridge contours from fingerprints. This work followed an isotropic model not considering the adaptive nature of real-life fingerprint samples. To address this problem, Zhao et al \cite{zhao2010adaptive} proposed an adaptive pore extraction model following a dynamic anisotropic model that estimates orientation and scale of pores dynamically. 

In \cite{zhao2009direct}, a direct pore matching approach is implemented making the process of pore matching independent from minutia matching. They used a pairwise pore comparison based on the corresponding pore descriptor that uses local features of pores. Afterwards, pore correspondences are refined using the RANSAC (Random Sample Consensus) algorithm to evaluate final pore matching result. A modified direct pore matching method is presented in the work of Liu et al. \cite{liu2010fingerprint} considering a sparse representation of finger pores that calculates the difference between pores to establish pore correspondences. It is then refined by a weighted RANSAC algorithm removing the false correspondences. Teixeira and Leite \cite{teixeira2014improving} proposed a method for pore extraction using spatial pore analysis that can accomodate varying ridge widths and pore sizes. Xu et al. \cite{xu2017fingerprint} proposed an approach that uses the size of the connected region of closed pores and the skeleton of valleys to detect open pores. 

Due to the excellent feature extraction capability of convolutional neural networks (CNNs), they have been used in recent years for pore extraction as a promising new approach. Su et al. \cite{su2017deep} first proposed a CNN-based approach for pore extraction showing comparable results with traditional approaches. Another pore extraction framework, DeepPore has been designed by Jang et al. \cite{jang2017deeppore} that uses pore intensity refinement to identify pores with higher true detection rate. The work in \cite{wang2017fingerprint} used the U-Net architecture \cite{ronneberger2015u} to extract ridges and pores present in fingerprints. Labati et al. \cite{labati2018novel} proposed a CNN-based pore extraction model that can handle heterogeneous fingerprint samples such as touch-based, touchless and latent fingerprints. Dahia et al. \cite{dahia2018improving} designed a fully convolutional network for pore detection from high resolution fingerprint images ensuring minimum number of required model parameters. 
  
Recently, Shen et al. \cite{shen2019stable} proposed a fully convolutional network incorporating focal loss \cite{lin2017focal} that solves the class imbalance problem. This method used an edge blur and shortcut structure that helps to utilize contextual information for pore detection. Nguyen et al. \cite{nguyen2019end} proposed a method for end-to end pore extraction for latent fingerprint matching where pore matching uses the ranked score information of  minutia matching. Anand et al. \cite{anand2020porenet} proposed a residual CNN based learning framework employing two models, DeepResPore  and PoreNet, for high resolution fingerprint images, surpassing the result of the state-of-the-art methods. Therefore, in this paper, the DeepResPore Network is used to detect pores utilizing pore intensity map while the PoreNet is trained to learn feature descriptors from pore patches by generating a deep feature embedding for corresponding fingerprint images. 
\subsection{Application of Super-Resolution on Fingerprint Image}
The impact of super-resolution (SR) on object recognition has recently received much attention from the research community; however, the impact of super-resolution on fingerprint recognition has not yet been thoroughly explored. Yuan et al. \cite{yuan2009fingerprint} considered SR as a pre-processing step for fingerprint image enhancement. This method applied early stopping, a regularization technique for improved image quality. A fingerprint SR image reconstruction approach applied in \cite{bian2016fingerprint} used sparse representation with a ridge pattern constraint. This method classified fingerprint patches considering ridge orientations and quality of the samples and learned coupled dictionaries for fingerprint classification. A ridge orientation-based clustering followed by sparse SR is adopted in the work of \cite{singh2015fingerprint}. This approach employs dominant orientation-based subdictionaries for sparse modeling of fingerprint data that significantly improved fingerprint recognition performance. Results reported in these studies demonstrated the use of SR as a promising scheme for fingerprint recognition, which is further investigated in this paper. 
\begin{figure*}
\centering
      \includegraphics[width=12cm,height=6cm]{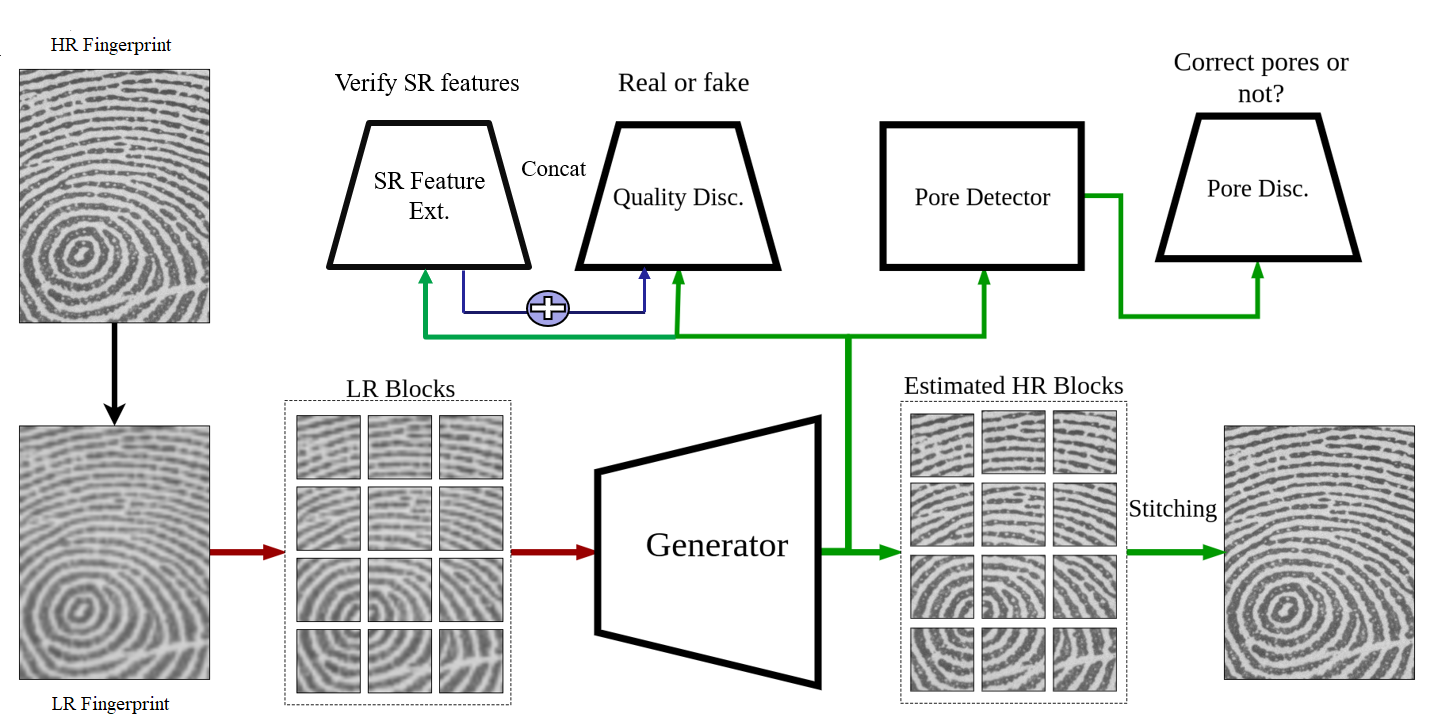}
  \caption{\label{fig:cdiag}Complete diagram of our proposed framework including the generator, quality discriminator, pore detector, pore discriminator and SR feature extractor.}
\end{figure*}

\section{Proposed Framework}

In this section, we present the details of our framework that is designed to convert a low-resolution (LR) fingerprint image into its high-resolution (HR) equivalent using a conditional Generative Adversarial Network (cGAN) architecture followed by a pore detector. 
\subsection{Conditional Generative Adversarial Network}
A GAN \cite{goodfellow2014generative} is a generative model that has outperformed other models in the task of synthetic image generation. This model has been explored in other representation learning tasks such as image super-resolution, image translation, etc. The conventional GAN model uses two sub-networks: a Generator, $(G)$ and a Discriminator, $(D)$. The generator tries to produce realistic samples by learning a mapping from a random noise z to an output y, such that  $y = G(z)$. Simultaneously, the discriminator learns to classify real and synthesized samples by distinguishing them. This system can be considered as a two-player min-max game where $G$ tries to fool $D$  by producing more samples indistinguishable from the real ones while $D$ gradually improves learning to detect the fake samples generated by $G$. The Conditional GAN (cGAN) \cite{isola2017image} differs from the conventional GAN \cite{goodfellow2014generative} in the sense that target sample along with noise are fed to the network such that $y = G(x,z)$; allowing target sample generation. The objective function for cGAN can be represented using the following equation:

\begin{equation}\label{eq:1}
\begin{split}
\min\limits_{G}\max\limits_{D} L_{GAN}(G,D) &= \min\limits_{G}\max\limits_{D}[\mathbb{E}_{x,y}[log D(x,y)]\\&\quad+\mathbb{E}_{x,z}[log(1-D(G(x,z)))].
\end{split}
\end{equation}

Here, the generator $G$ constantly attempts to minimize Eq. \ref{eq:1} and the discriminator $D$ tries to maximize it. 

\subsection{Training Objective}

The goal of this paper is to design an efficient fingerprint SR model that is guided by a finger pore detection model in order to improve the overall fingerprint recognition accuracy. The network is trained in an end-to-end fashion such that the two individual models, the super-resolution and pore detector get benefit from each other. To achieve stable convergence of the model, we incorporate three losses from the SRGAN model: the Mean Squared Error (MSE) loss, adversarial loss \cite{goodfellow2014generative}, and perceptual loss \cite{johnson2016perceptual}. In addition, to preserve class identification details embedded in ridge patterns and pores of fingerprints, we add two more losses to the model. One is the ridge reconstruction loss that considers ridge pattern variations and the other one is the pore detection loss, which uses a pore location map to identify pore map differences between the ground truth and the super-resolved fingerprint samples. 

\subsubsection{MSE Loss}
Most state-of-the-art SR methods \cite{shi2016real}\cite{dong2015image}
use MSE loss, as this helps to achieve high peak signal-to-noise ratio. Similar to those methods, we also used MSE loss in our model. This loss estimates content-wise dissimilarity by taking the absolute pixel-wise differences between the generated image and the ground truth, as given by:
\begin{equation}
l_{MSE} = \frac{1}{N}\sum\limits_{n=1}^N\frac{1}{WH}\sum\limits_{w=1}^W \sum\limits_{h=1}^H \norm{(I_{n}^{HR})_{w,h}-G(I_{n}^{LR})_{w,h}}^2,
\end{equation}where W and H represent width and height of a fingerprint sample and N is the number of training samples. 
From Eq. 4, we see that the loss is simply the $L2$ difference between the ground truth HR image $I_{n}^{HR}$ and the generated super-resolved image $G(I_{n}^{LR})$ from the LR image $I_{n}^{LR}$.
\subsubsection{Adversarial Loss}
The generator in our model uses the adversarial loss that aids in the generation of natural looking images. The concept is that the discriminator tries to maximize the probability of genuine or fake images calculated from the real images and the probability of the fake ones denoted by $log(1-D(I^{LR},G(I^{LR}))$ while the generator tries to minimize the prediction of the fake samples by the discriminator; thus, promoting a lower chance of fake sample generation. We can mathematically formulate this loss as follows:
\begin{equation}\begin{split}\label{eq:4}
\ell_{adv} &= \min_{G}\max_{D}[E_{I^{HR}\sim P_{train}(I^{HR})}[log D(I^{LR},I^{HR})]\quad \\& + E_{I^{LR}\sim P_{G}(I^{LR})}[log(1-D(I^{LR},G(I^{LR})))],
\end{split}
\end{equation}where $P_{train}(I^{HR})$ and $P_{G}(I^{LR})$ denote the probability distributions of real high-resolution fingerprint images and the corresponding low-resolution fingerprint images, respectively. 
\subsubsection{Perceptual Loss}
To preserve the inherent details of a ground truth in the generated fingerprint image, we use the perceptual loss proposed by Ledig et al. \cite{johnson2016perceptual}. A pretrained 19- layer VGG network \cite{simonyan2018very} is used to extract abstract features from images that preserve the discriminative information of the images in a lower dimensional sub-space. The perceptual loss is the L2 distance between the ground truth and the super-resolved image, which is measured as: 
\begin{equation}
\begin{split}
\ell_{per}^{SR} = \frac{1}{N}\sum\limits_{n=1}^N\frac{1}{C_{j}W_{j}H_{j}}\sum\limits_{c=1}^{C_{j}}\sum\limits_{w=1}^{W_{j}}\sum\limits_{h=1}^{H_{j}}||\phi_{j}^{c}(I_{n}^{HR})_{w,h}\\-\phi_{j}^{c}(G(I_{n}^{LR})_{w,h}||^2,
\end{split}
\end{equation}
where $\phi_{j}^{c}$ represents the $c^{th}$ feature map in the   $\textit{j}^{th}$ convolutional layer and $C_{j}$, ${W_{j}}$ and ${H_{j}}$ denote the $j^{th}$ layer dimension.

\subsubsection{Ridge Reconstruction Loss}
Similar to \cite{johnson2016perceptual}, the ridge reconstruction loss is computed
as the squared Frobenius norm of the difference between the Gram matrices of the output and target images. The Gram matrix $G^\phi_i (y)$  defines the style of the ground truth image by computing every feature activation of entire ground truth feature map space. Mathematically, it can be expressed as the matrix multiplication of each of the activation and the transpose of feature activation matrix whose elements can be obtained using Eq. \ref{eq:gram} which is given by: 
\begin{equation}
\label{eq:gram}
\begin{split}
G_{j}^{\phi}(y)_{c,c'} = \frac{1}{N}\sum\limits_{n=1}^N\frac{1}{C_{j}W_{j}H_{j}}\sum\limits_{w=1}^{W_{j}}\sum\limits_{h=1}^{H_{j}}\phi_{j}(y_{n})_{c,w,h}\\\phi_{j}(y_{n})_{c',w,h},
\end{split}
\end{equation}
where $\phi_{j}$ represents the activations of the $\textit{j}^{th}$ convolutional layer.
For the generated image $\tilde{y}$, we calculate the Gram matrix in a similar fashion which is given by:
\begin{equation}
\label{eq:gram1}
\begin{split}
G_{j}^{\phi}(\tilde{y})_{c,c'} = \frac{1}{N}\sum\limits_{n=1}^N\frac{1}{C_{j}W_{j}H_{j}}\sum\limits_{w=1}^{W_{j}}\sum\limits_{h=1}^{H_{j}}\phi_{j}(\tilde{y_{n}})_{c,w,h}\\\phi_{j}(\tilde{y_{n}})_{c',w,h}.
\end{split}
\end{equation}
Finally, the ridge reconstruction loss of the network is given by:
\begin{equation}\label{eq:ridge}
\ell_{ridge}^{J}= \sum\limits_{j=1}^J||G_{j}^{\phi}(\tilde{y})-G_{j}^{\phi}(y)||_F^2.
\end{equation}

\subsubsection{Pore Detection Loss}
 Let $I_{n}^{LR}$ be a LR fingerprint that is to be translated into the HR space represented by $\tilde{I}_{n}^{HR}$ which is close to the original high-resolution fingerprint $I_{n}^{HR}$. The pore detection module $P(.)$ utilizes the generated $\tilde{I}_{n}^{HR}$ and produces the pore intensity map, which is compared with the original pore map, $p_h$. To minimize the error in each iteration, the L1 error between the two intensity maps is back-propagated. The loss due to the pore detector can be expressed as:
\begin{equation}\label{eq:5}
\ell_{pore}= \frac{1}{N}\sum\limits_{n=1}^N||P(G(I_n^{LR})-p_h)||_1,
\end{equation}
where $P(G(I_n^{LR}))$ estimates the pore locations denoted by marked pores for the input fingerprint. 
\subsubsection{Final Loss Function}
The overall loss for the model can be written as the combination of all the above-mentioned losses with appropriate weighting which is formulated below as: 
\begin{equation}
\begin{split}
\ell_{Total}^{G} = \lambda_1\ell_{MSE} + \lambda_2  \ell_{adv}+\lambda_3\ell_{per}\\
    +\lambda_4\ell_{ridge} +\lambda_5 \ell_{pore},
\end{split}
\end{equation} 
where $\lambda_1$, $\lambda_2$, $\lambda_3$, $\lambda_4$ and $\lambda_5$ are used as the constraints to balance the associated losses.
The combination of $\ell_{MSE}$, $\ell_{adv}$ and $\ell_{per}$ losses lead to the generation of realistic fingerprint images. Setting $\lambda_1$, $\lambda_2$, $\lambda_3$ to $10^{-3}$ as weighting factors helps these losses to converge faster. The ridge reconstruction loss, $\ell_{ridge}$  enables the model to transfer the correct ridge patterns to the generated fingerprint sample and the pore detection loss, $\ell_{pore}$ adds pore details to the generated fingerprint samples to ensure a high-performance fingerprint matcher. From our empirical study, we have set the value of $\lambda_4$, $\lambda_5$ to $10^{-2}$ to achieve the best performing model.
\subsection{Network Architecture}

We have combined three separate models into one to create a joint model that is able to produce a high-resolution fingerprint from a low resolution one, as shown in Fig. \ref{fig:cdiag}.   
\subsubsection{Super-Resolution Model}
The first model in our network is the super-resolution model. The task of this model is to predict a high-resolution fingerprint image from its low-resolution version. We have designed this model with inspiration from \cite{ledig2017photo}. Our generator network has seven residual blocks that have identical layout. Each residual block has two convolutional layers of 3$\times$3 kernel with stride 1, 64 feature maps followed by batch-normalization layer \cite{ioffe2015batch}, and ParametricRELU \cite{he2015delving} as the activation function. Two sub-pixel convolutional layers \cite{shi2016real} are attached to the network to produce a high-resolution image from its low-resolution version.
Our quality discriminator adopts a design similar to the guidelines from Radford et al. \cite{radford2015unsupervised}. It has seven convolutional layers each having 3$\times$3 filter kernels. As the network advances, image resolution is decreased by strided convolution with increasing feature map size. The leakyRELU activation function is used with $\alpha=0.2$ in this network. Two dense layers and a sigmoid function are added so that the network can distinguish between real and generated samples. 

\subsubsection{Deep ID Extractor}
 In order to preserve class identity information in our model, we employ a deep Siamese verifier as a feature extractor \cite{dabouei2018id}. First, we train the verifier with low-resolution fingerprint samples using the contrastive loss \cite{chopra2005learning}. Then, we extract features from the generated samples using this pre-trained module. To make sure the discriminator is considering the identity information, feature maps from the first, second and third layer of the verifier (size: 40x30x64, 20x15x128 and 10x8x256) are concatenated depth-wise with the features from the quality discriminator making the final output feature map size 40x30x128, 20x15x256 and 10x8x512 respectively. All of the layers  comprise of convolutional layers and LeakyReLU activation functions. Kernel size for all the convolutional layer is 3$\times$3 and stride is set to two. 

\subsubsection{Pore Detection Model}
The pore detection model \cite{anand2019pore} follows a residual structure. The network has eight residual blocks with eight shortcut connections. All the residual blocks use a 3$\times$3 kernel with depth increasing by a factor of two. In total, the network is comprised of eighteen layers with 1$\times$1 convolution and shortcut connections arranged in alternating manner. The deep residual network takes an input patch size of 80$\times$60 and generates a similar size pore intensity map with marked pores. Then, a binarized pore map is created to highlight the position of the pores.
\begin{figure*}
\centering
      \includegraphics[width=15cm,height=7cm]{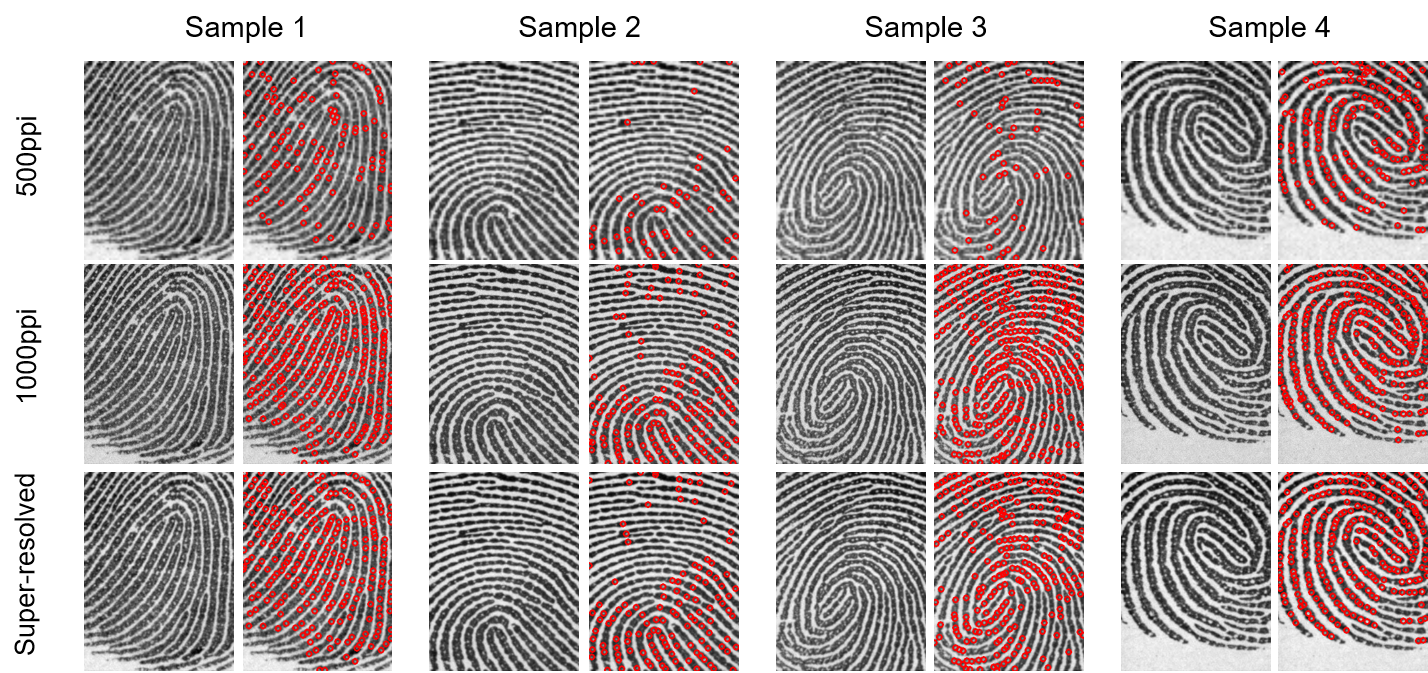}
  \caption{Pore detection results in LR, corresponding real HR and super-resolved samples generated by our model. Here, red circles represent detected pores.}
\end{figure*}

\section{Experiments and Result Analysis}
In this section, we present our experimental results to analyze the impact of super resolution on finger pore detection. We conduct our experiment on two publicly available datasets (PolyU HRF DBI \cite{polyU} and FVC2000 DB1 \cite{maio2002fvc2000}). The PolyU HRF DBI dataset has images of 1,200 ppi resolution with spatial size of 320$\times$240. The annotations provided with this dataset contain the pore locations denoted as the central coordinates of pores. The dataset has 30 annotated fingerprint images. We use patches extracted from the first 20 for training  and the remaining to test the performance of the pore detector. We have also applied data augmentation to increase the number of samples in our dataset. We augmented our train set by applying gamma transformation, random scale, horizontal and vertical flip to create different contrast for the original samples. To further increase the size of the training set, we divide the images into overlapping patches of size 40$\times$30 and feed to the generator. The synthesized super-resolved patches of size 80$\times$60 are used to train the pore detector. The second dataset FVC2000 DB1 has images of 500 ppi resolution with spatial size of 300$\times$300. Both PolyU DBI-test and FVC2000 DB1 are used to evaluate the performance of our proposed approach.

\subsubsection{Training}
First, we train our super-resolution and pore detection network separately. For training the super-resolution network, we use the Adam optimizer with a momentum of 0.9, beta 0.5 and batch size of 64. The learning rate is set as $10^{-4}$. Our pore detector has been trained using the Adam optimizer with a batch size of 64 and a learning rate of $10^{-3}$ for 30 epochs. For joint training of our model, the entire model is trained with pre-trained weights of  super-resolution and pore detection networks for 20 epochs and then the weights are updated for 20 epochs. 
\subsubsection{Quality distribution analysis}

To evaluate the quality of the super-resolved samples, we have used the NFIQ 2.0 utility from NBIS \cite{tabassi2004nist}. The NFIQ 2.0 assigns a quality score to an image ranging from 0 to 100. From the quality score distribution in Fig. 4, a large overlap among the scores of the generated and real HR fingerprints is visible which indicates the qualitative similarity of the generated fingerprint samples of our model to the real HR fingerprint samples. Approximately 79\% of the generated samples have been assigned a quality score of 50 or higher which confirms quality image generation by our modified SRGAN.
\begin{figure}[t]
\centering
      \includegraphics[width=8cm, height=4.5cm]{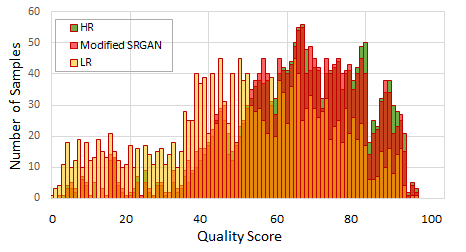}
  \caption{\label{fig: nf1}Image quality comparison of LR, real HR and generated samples from our modified SRGAN for an upscale factor 2$\times$.}
\end{figure}

\subsubsection{Performance analysis of pore detection}
The pore detection performance of our proposed method is demonstrated in Fig. 3. We have also summarized our pore detection model performance with other state-of-the-art methods in Table \ref{tab:tdr}. This experimental result is reported considering the True Detection Rate (TDR)  and False Detection Rate (FDR) of pores in the PolyU DBI dataset.  
 
From Table \ref{tab:tdr}, we can conclude that our pore detection model performs significantly higher, surpassing the results of other baseline methods. The proposed pore detection method achieves a high TDR accompanied by a low FDR, yielding a very high accuracy. 

\begin{table}[b]
\caption{\label{tab:tdr} Pore detection performance comparison in terms of TDR and FDR. Results reported apart from ours have been collected from the corresponding papers.}
\begin{center}
\begin{tabular}{l|c|c}
\hline
Method & TDR  & FDR  \\
\hline
\hline
Jain et al. \cite{jain2006pores} & 75.9\% & 23\% \\
\hline
Zhao et al. \cite{zhao2010adaptive} & 84.8\% & 17.6\% \\
\hline
Segundo et al. \cite{pamplona2015pore} & 90.8\% & 11.1\% \\
\hline
Su et al. \cite{su2017deep} & 88.6\% & 0.4\% \\
\hline
Dahia et al. \cite{dahia2018improving} & 91.95\%
 & 8.88\%\\
\hline
Ours & \textbf{94.3}\% & \textbf{0.4}\% \\
\hline

\end{tabular}
\end{center}
\end{table}

\subsubsection{Performance analysis of unified model}
In this work, we have used all three levels of fingerprint features. Ridge patterns and minutiae from fingerprints are extracted by applying wavelet-based Gabor filtering \cite{zhang2002wavelet} and crossing number algorithm \cite{mehtre1993fingerprint}, respectively. The matching at Level-2 is performed combining two different matchers, namely correlation-based \cite{jain1999combining} and minutiae-based matchers \cite{jain1999combining}. To compare fingerprints based on pores, Graph Comparison Algorithm \cite{xu2018fingerprint} is utilized which focuses on local features and spatial relationship between pores. In order to make use of the extended fingerprint feature set, a score-level fusion of match scores from Level 1, Level 2 and Level 3 features is performed using sum-rule and min-max-normalization \cite{ross2003information} to conduct fingerprint matching. 

\begin{figure}
\centering
      \includegraphics[width=8cm, height=5.5cm,scale=0.45]{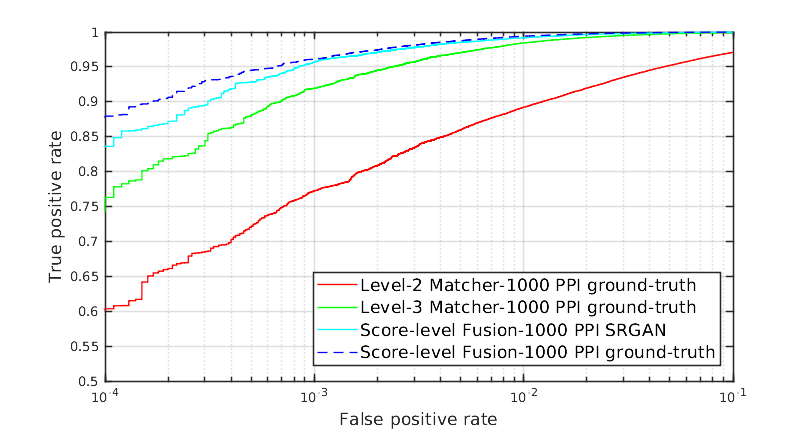}
  \caption{\label{fig: polyu} ROC curves for fingerprint recognition using the different level features extracted from the PolyU HRF DBI dataset.}
\end{figure}

To evaluate the fingerprint recognition accuracy of our approach, we have generated 3,700 genuine pairs and 21,756 imposter pairs following the same procedure in \cite{liu2011novel,liu2010fingerprint} from the PolyU dataset. The genuine pairs are obtained by matching each fingerprint image in the first session with all five images of the same finger in the second session. The first fingerprint image of each finger in the second session is compared with the first fingerprint image of all the other fingers in the first session. We compare our method with other existing methods; MICPP \cite{jain2006pores}, MINU\_SRDP \cite{liu2010fingerprint}, TDSWR \cite{liu2011novel}. Based on the performed matching experiments, we find that the performance of matching 1000 ppi is quite the same as matching 1200 ppi samples. Hence, we select 1000 ppi as our preferred resolution. It can be observed from Table \ref{tab:real} that Equal Error Rate (EER) of our SR fingerprints is comparable to that of the ground truth fingerprint images, indicating the effectiveness of our proposed SR based pore detection method. In order to analyze the impact of different level fingerprint features in recognition performance, we have plotted the ROC for Level-2, Level-3 and the score-level fusion of the match scores for both matchers. From Fig. \ref{fig: polyu}, we see that combination of match scores from Level-2 and Level-3 shows significant performance gain compared to individual Level-2 and Level-3 matchers. Also, it can be observed that the recognition accuracy of generated fingerprint samples is very close to the accuracy obtained using real ground-truth samples. 

\begin{table}[b]
\caption{\label{tab:real} Fingerprint recognition  performance comparison of different pore matching methods in terms of EER.}

\begin{center}
\begin{tabular}{l|c}
\hline
Method & EER \\
\hline
\hline
MICPP &30.45\% \\
\hline
MINU\_SRDP & 5.41\% \\
\hline
TDSWR & 3.25\% \\
\hline
Ours (1000-ppi Ground Truth) & \textbf{1.57\%} \\ 
\hline
Ours (1000-ppi Modified SRGAN) & \textbf{1.63\%} \\ 
\hline
\end{tabular}
\end{center}
\end{table}

\begin{figure}
\centering
     \includegraphics[width=7.5cm, height=5cm,scale=0.45]{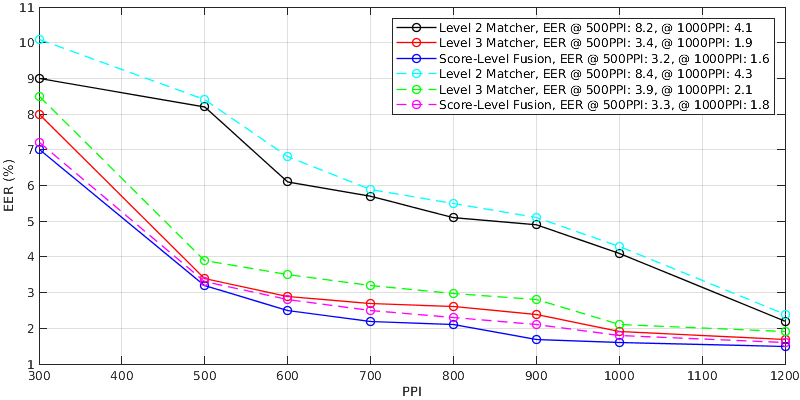}
  \caption{\label{fig: eer_all} Fingerprint recognition performance evaluated at multiple resolutions using different level features for an upscale factor 2$\times$. Solid lines represent the EER values for the PolyU DBI dataset and dashed lines are for the EER values in the FVC2000 DB1 dataset.}
\end{figure}

Fig. \ref{fig: eer_all}. shows the fingerprint recognition performance of our model across different image resolutions for both the PolyU DBI and the FVC2000 DB1 dataset. It can be observed that the decrease in EER is almost twice for the super-resolved high-resolution samples (e.g., 1000 ppi) from their ground-truth (e.g., 500 ppi) samples. This ensures reliable reconstruction of HR fingerprint samples using our model. The significant decrease in EER over the two datasets show a consistent improvement in fingerprint recognition using our SR guided joint framework. It is demonstrated that the generated 1000 ppi fingerprints can substantially improve the matching performance compared to the ground-truth 500 ppi samples. 


\begin{figure}[b]
\centering
    \includegraphics[width=9 cm, height=6 cm,scale=0.25]{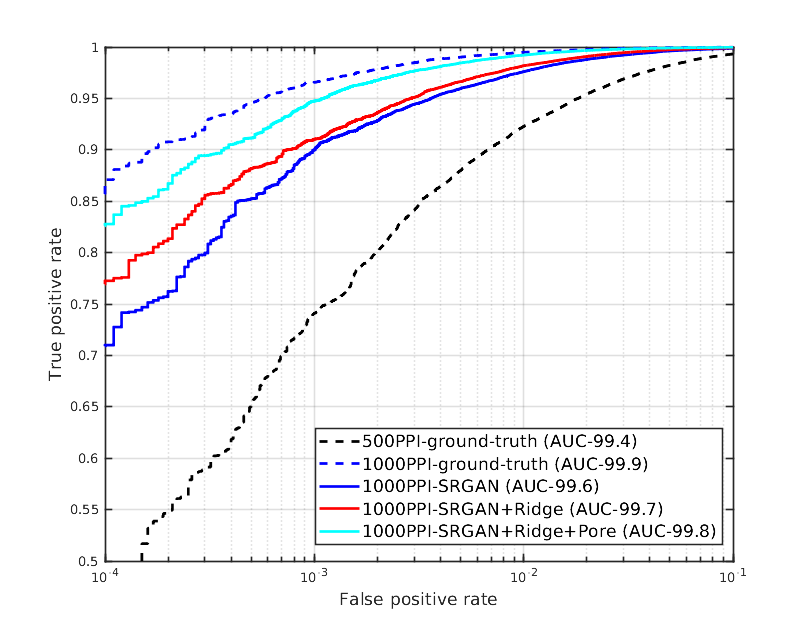}
  \caption{\label{fig:frep}ROC curves for real HR, generated samples from SRGAN and our modified SRGAN for an upscale factor 2$\times$.}
\end{figure}

To provide a comprehensive performance analysis, we plot ROC curves investigating the effect of different loss functions adopted in our approach. In Fig. \ref{fig:frep}, we can see that the matching results based on the SRGAN losses combined with the ridge and pore detection losses achieve the highest Area Under the Curve (AUC) of around 99.8\%, which is very close to the AUC computed from the ground truth HR fingerprint samples. It clearly demonstrates that introducing ridge reconstruction and pore detection losses help the model to generate samples close to real ones, which provides an overall improvement in fingerprint recognition performance. 

\section{Conclusion}
This paper proposes a jointly optimized fingerprint recognition framework using the concept of super-resolution and pore detection. The model is able to generate HR fingerprint samples, learn pore locations, ridge structure and other details from LR samples. The increase in resolution helps to achieve a high pore detection accuracy, which in turn forces the generator to produce high quality synthesized fingerprint samples. Also, integrating features extracted from a deep verifier with a quality discriminator preserves the individuality in our reconstructed samples. Reliable reconstruction of 1000 ppi fingerprint from its 500 ppi equivalent proves the validity of our approach.

{\small
\bibliographystyle{IEEEtranS}
\bibliography{bare_conf}
}

\end{document}